# Imaging-based histological features are predictive of *MET* alterations in Non-Small Cell Lung Cancer

## Authors & Affiliations

**Authors:**
Rohan P Joshi[1,*]
Bolesław L Osinski[1]
Niha Beig[1]
Lingdao Sha[2]
Kshitij Ingale[1]
Martin C Stumpe[1]

**Affiliations:**
1. Tempus Labs Inc., Chicago IL 60654
2. Work done at Tempus Labs Inc, current affiliation: Amazon
* Address correspondence to: rohan.joshi@tempus.com

## Abstract

*MET* is a proto-oncogene whose somatic activation in non-small cell lung cancer leads to increased cell growth and tumor progression. The two major classes of *MET* alterations are gene amplification and exon 14 deletion, both of which are therapeutic targets and detectable using existing molecular assays. However, existing tests are limited by their consumption of valuable tissue, cost and complexity that prevent widespread use. *MET* alterations could have an effect on cell morphology, and quantifying these associations could open new avenues for research and development of morphology-based screening tools. Using H&E-stained whole slide images (WSIs), we investigated the association of distinct cell-morphological features with *MET* amplifications and *MET* exon 14 deletions. We found that cell shape, color, grayscale intensity, and texture-based features from both tumor infiltrating lymphocytes and tumor cells distinguished *MET* wild-type from *MET* amplified or *MET* exon 14 deletion cases. The association of individual cell features with *MET* alterations suggested a predictive model could distinguish *MET* wild-type from *MET* amplification or *MET* exon 14 deletion. We therefore developed an L1-penalized logistic regression model, achieving a mean Area Under the Receiver Operating Characteristic Curve (ROC-AUC) of 0.77 ± 0.05sd in cross-validation and 0.77 on an independent holdout test set. A sparse set of 43 features differentiated these classes, which included features similar to what was found in the univariate analysis as well as the percent of tumor cells in the tissue. Our study demonstrates that *MET* alterations result in a detectable morphological signal in tumor cells and lymphocytes. These results suggest that development of low-cost predictive models based on H&E-stained WSIs may improve screening for *MET* altered tumors.

## Introduction

Non-small cell lung cancer (NSCLC) is the second most common cause of cancer worldwide and one of the leading causes of death in the United States (Sung et al. 2021). Over the past few decades, therapy for NSCLC has been at the forefront of the precision medicine era, starting with therapies targeting specific mutations in the gene *EGFR* (Lynch et al. 2004). Today, precision medicine in NSCLC has matured. Drugs are available to target numerous additional genes, including *ALK, ROS1, ERBB2, KRAS, RET,* and *BRAF*, and gene alterations are routinely detected using high throughput assays like next-generation sequencing (NGS) (Planchard et al. 2018; Jiang et al. 2018; Yang et al. 2020). These mutation-targeting therapies have revolutionized patient care and often provide patients with longer survival and fewer side effects compared to chemotherapy.

In recent years, *MET* has emerged as a promising therapeutic target in NSCLC. *MET* is a proto-oncogene whose constitutive activation leads to cell proliferation, tumor development and tumor progression (Organ and Tsao 2011; Zhang et al. 2018). Constitutive activation of MET is therefore associated with poor prognosis (Awad et al. 2016). Somatic mutations and amplifications are the two major classes of alterations that lead to constitutive *MET* activation. Somatic mutation predominantly causes exon 14 deletion—which alters degradation of the resultant protein—and *MET* amplification results in increased protein expression (Drilon et al. 2017; Peng et al. 2021; Heydt et al. 2019). *MET* amplification is one of the mechanisms that confers resistance to EGFR inhibitors such as osimertinib in NSCLC (Engelman et al. 2007; Leonetti et al. 2019). As a tyrosine kinase involved both in primary cancer and therapeutic resistance, MET is an attractive therapeutic target. MET inhibitors such as capmatinib and tepotinib have demonstrated efficacy in NSCLC patients harboring a *MET* exon 14 deletions (Mathieu et al. 2022). In addition, ongoing clinical trials evaluating the efficacy of inhibitors in *MET* amplified NSCLC with and without EGFR inhibitor resistance have shown promising initial results (Camidge et al. 2021; Park et al. 2021; Wolf et al. 2020).

Given the occurrence of both expression-related and protein coding changes in MET, multiple techniques such as NGS, fluorescence *in situ* hybridization (FISH), and immunohistochemistry (IHC) are used to identify various aspects of *MET* status (Fang et al. 2018; Peng et al. 2021). MET alterations could also trigger biological processes that result in changes in cellular morphology, potentially allowing for detection using routinely available hematoxylin and eosin (H&E) stained slides. Application of computer vision to digital pathology has enabled quantification of morphologic phenotypes previously unrecognized by pathologists, including phenotypes related to gene mutations (Coudray et al. 2018). In addition, interpretable machine learning models can be used to generate testable hypotheses to better understand tumor biology in a manner that is complementary to molecular assays (Diao et al. 2021; Bera et al. 2019).

In this work, we identify the presence of a morphological signal from immune and tumor cells that can be attributed to *MET* alterations. We further present an H&E imaging-based model to predict *MET* genomic alterations (exon 14 deletion and amplification). With further validation of this imaging phenotype, low-cost predictive models based on H&E-stained WSIs could be created to provide further insight into tumor biology and ultimately aid in drug development.

## Methods

### Dataset - H&E WSIs sample preparation and digitization

Archival formalin-fixed, paraffin-embedded (FFPE) tumor tissues (1 tissue sample/patient) from NSCLC patients were used in this study. All the FFPE blocks were processed, reviewed, and stored in a College of American Pathologists (CAP) accredited and Clinical Laboratory Improvement Amendments (CLIA) certified laboratory (Tempus Labs Inc., Chicago, IL, USA).

Each FFPE block was cut into 4 µm-thick serial sections and affixed to glass slides. One of the sections was baked in a Premier Scientific Slide Warmer and stained for H&E staining using a Tissue Tek Prisma + (Sakura, Torrance, CA) automatic stainer. The slide was then reviewed by board-certified pathologists. Slides were scanned at 40X in accordance with manufacturer's instructions for Philips IntelliSite Ultra Fast Scanner (UFS) and stripped of any patient-identifying information.

### Dataset - assessment of *MET* alterations (amplification and exon 14 deletion)

The current reference method for assessment of *MET* amplification is FISH, which allows a per tumor cell direct examination of the number of *MET* copies and—with the addition of a centromeric marker—the relative number of *MET* copies compared to the number of copies of chromosomes on which *MET* resides. FISH thresholds for *MET* amplification include copy number (CN) >= 5 or *MET:CEP7* ratio >= 2, which stratify patients into different survival categories (Peng et al. 2021; Fang et al. 2018). Similar FISH thresholds have been used as enrollment criteria in clinical trials (Cheng et al. 2018; Song et al. 2019; Wu et al. 2018). The Tempus xT assay reports CN = 0 as a loss, and CN > 8 as a gain (where a copy number of 8 means 8 DNA copies of a gene were identified per tumor cell). However, in this feasibility study we defined CN >= 5 as a *MET* amplification based on previous studies using FISH. To this end, NGS was performed via the Tempus xT assay, encompassing DNA-seq of 595-648 genes as well as whole exome capture RNA-seq, as described previously (Beaubier et al. 2018; 2019; Hu et al. 2021).

*MET* exon 14 deletion events were identified by expert curation of DNA sequencing and include genomic variants that alter a splicing site or delete the whole exon. For a minority of samples, RNA sequencing was available and the direct result of altered splicing (observed as "fusion" of exon 13 to 15) was further required to confirm the exon 14 deletion event (Beaubier et al. 2019).

### Dataset - inclusion and exclusion criteria

The following inclusion criteria were used for this study: i) availability of H&E-stained WSIs obtained at a magnification of 40X (0.25 µm per pixel) on a Philips UFS, and ii) availability of pathology report confirming primary NSCLC or metastatic NSCLC diagnosis within the lung (either through biopsy, surgical resection, core needle biopsy, or transbronchial biopsy). In all cases, images with altered spatial architecture and/or cell morphology due to retrieval procedures (such as fluid aspirate or fine needle aspirate cases, cytology or flash frozen cases) were excluded.

The following additional exclusion criteria were used to remove images from further analyses: i) slides with less than 100 viable tumor cells, ii) slides with more than 60% of the cancer area out of focus, or iii) if the stained tissue had an unexpected color for H&E (either too faded or not typical H&E color patterns) as identified by a board-certified pathologist. Slides with artifacts such as pen marks, tissue folding, bubbles, crush, shatter present within more than 50% of the tumor region of the tissue as identified by Tempus Labs Inc. proprietary automated algorithms were also removed.

While curating the ground truth, NSCLC cases harboring *MET* CNs of 5, 6, or 7 were manually reviewed by a pathologist to verify the gene segments involved in making the bioinformatics CN call. Cases with less than four consecutive gene segments with CN>=5 were excluded from the study. Further, we observed that for some cases with DNA CN=2, RNA expression of *MET* was high. Because the H&E image phenotype may be associated with DNA *MET* alterations, RNA expression, or both, the molecular phenotype of these cases are potentially noisy. Cases with *MET* expression greater than the 90th percentile (CN=2) were excluded from further analyses (note that this exclusion criteria applies only to the NSCLC samples with sufficient tissue for RNA sequencing).

| **Characteristic** | **Training Cohort (n=528)** | | | **Holdout Test Cohort (n=176)** | | |
|---|---|---|---|---|---|---|
| | ***MET* Wild-Type** | ***MET* altered** | **p-value** | ***MET* Wild-Type** | ***MET* altered** | **p-value** |
| ***MET* alteration** | | | | | | |
| **Wild-Type** | 425 | 0 | - | 141 | 0 | - |
| **CNV >= 5** | 0 | 61 | | 0 | 15 | |
| **Exon 14 deletion** | 0 | 39 | | 0 | 20 | |
| **CNV >= 5 and Exon 14 deletion** | 0 | 3 | | 0 | 0 | |
| ***Histology*** | | | | | | |
| **Adenocarcinoma** | 246 | 69 | *0.454* | 77 | 25 | *0.455* |
| **Squamous** | 76 | 15 | | 27 | 5 | |
| **Adenosquamous** | 13 | 1 | | 5 | 1 | |
| **Other** | 12 | 2 | | 5 | 1 | |
| **Poorly differentiated** | 78 | 16 | | 27 | 3 | |
| ***Procedure Type*** | | | | | | |
| **Biopsy** | 23 | 6 | *0.939* | 8 | 2 | *0.998* |
| **Core Needle Biopsy** | 176 | 45 | | 58 | 15 | |
| **Excisional Biopsy** | 14 | 2 | | 5 | 1 | |
| **Resection** | 169 | 41 | | 56 | 14 | |
| **Transbronchial biopsy** | 43 | 9 | | 14 | 3 | |
| **Total** | **425** | **103** | | **141** | **35** | |

**Table 1**: Clinical characteristics of training and holdout test cohorts employed in our study. Significance testing using chi-squared tests between *MET* status for histology subtype and procedure type in both the training and independent holdout test cohort was found to be statistically not significant.

### Data preprocessing - tissue classification

Tumor segmentation was based on a custom multi field-of-view network with a fully convolutional ResNet-18 backbone pretrained with ImageNet weights (Sha et al. 2019). The network was trained on 77 NSCLC slides and validated on 11 holdout slides. Ground truth tumor regions on these slides were manually annotated by American Board of Pathology-certified pathologists using publicly available digital pathology software: QuPath (Bankhead et al. 2017). Annotation coverage of each slide was between 50–80%. Annotated regions were tiled into 1024x1024 tiles at 20X magnification, producing > 500,000 training set tiles; 80% of these were used for training and 20% were used for validation. Geometric and color-space image augmentations were performed batchwise during training. To improve regularization and accelerate training, batch normalization was implemented. Cross entropy loss was minimized using the Adam optimizer and an initial learning rate 1e-3, which was decreased by 0.5 every 2 epochs. The models were trained until the validation loss no longer decreased, which occurred in the 10th epoch. The tumor classification ROC-AUC on the holdout test set was 0.947.

### Data pre-processing - cell detection

The cell segmentation model, based on the U-Net architecture (Ronneberger, Fischer, and Brox 2015), generated segmentations of individual tumor nuclei and lymphocytes throughout the whole image. For this purpose, we trained 2 separate models: i) a lymphocyte detector trained on annotations of only lymphocytes and ii) a general nucleus detector trained on annotations of all nuclei (including lymphocytes, tumor cells, fibroblasts, and epithelial cells). The lymphocyte detection training set consisted of 369 regions-of-interest (ROIs) sampled from 12 in-house scanned WSIs. The general nucleus detector training set consisted of 60 ROIs; 30 which were sampled from one in-house scanned slide and 30 sampled from TCGA diagnostic slides (one ROI per slide). Each ROI was 512 x 512 pixels at 20x magnification and was annotated in QuPath software (Bankhead et al. 2017) by board-certified pathologists who manually outlined every nucleus. Model selection was performed by selecting the model with highest DICE score (Dice 1945) on an independent holdout test set consisting of 100 pathologist annotated 512x512 ROIs from 50 slides.

The tissue and cell segmentation models are combined to assign identities to tumor and lymphocyte cells within a given region of interest. In this work, we only looked within the tumor regions to identify tumor cells and lymphocytes. Cells within an identified tumor tile were given one of two classes based on the following conditions: i) if a cell is detected by the lymphocyte segmentation model then it is classified as a lymphocyte, else ii) if a cell is detected by the general cell segmentation model but not by the lymphocyte segmentation model, then it is classified as a tumor cell of the tile.

### Feature extraction

A total of 372 cell-based features were extracted per slide, corresponding to one of five categories: cell nuclei percentage, shape, RGB color, grayscale intensity, and texture (for details see Table S1). All features were computed for tumor cells as well as Tumor Infiltrating Lymphocytes (TILs). Previous work has shown that the percentage of tumor cells and TILs correlates with outcomes across a wide variety of cancers (Smits et al. 2014of; Althammer et al. 2019; Li et al. 2021; Corredor et al. 2019). All non percentage features were restricted to tumor cells and lymphocytes within the tumor region identified by our tumor classification model. The shape features reflect nuclear size and shape variations, especially common in pleomorphic tumor cells, which may impact cell migration and proliferation (Bussolati et al. 2008; Lu et al. 2018; Tollis et al. 2022; Jevtić et al. 2014; Mukherjee et al. 2016). Color, grayscale intensity, and texture features were computed from bounding boxes surrounding each cell nucleus. The texture features were composed from a subset of Haralick features (R.M. Haralick 1979) and Grayscale Gradient features (Zwillinger and Kokoska 1999). Such cell texture features have been shown to distinguish between histological subtypes and stratify prognosis in NSCLC (Alvarez-Jimenez et al. 2020; Yu et al. 2016).

Shape, color, intensity, and texture features were computed for 20% of all lymphocytes and tumor cells in the tumor region to improve computational efficiency. Statistics such as mean, standard deviation, skewness and kurtosis values of cell-level features were then calculated and concatenated to generate slide level imaging features. Clipping (1–99 percentile) was implemented in this feature generation pipeline in order to suppress statistical outliers due to possible artifactual cell segmentations. Haralick features were extracted using the computer vision library for Python, Mahatos (Coelho 2013), and shape features were extracted using the OpenCV-Python library (Bradski 2000).

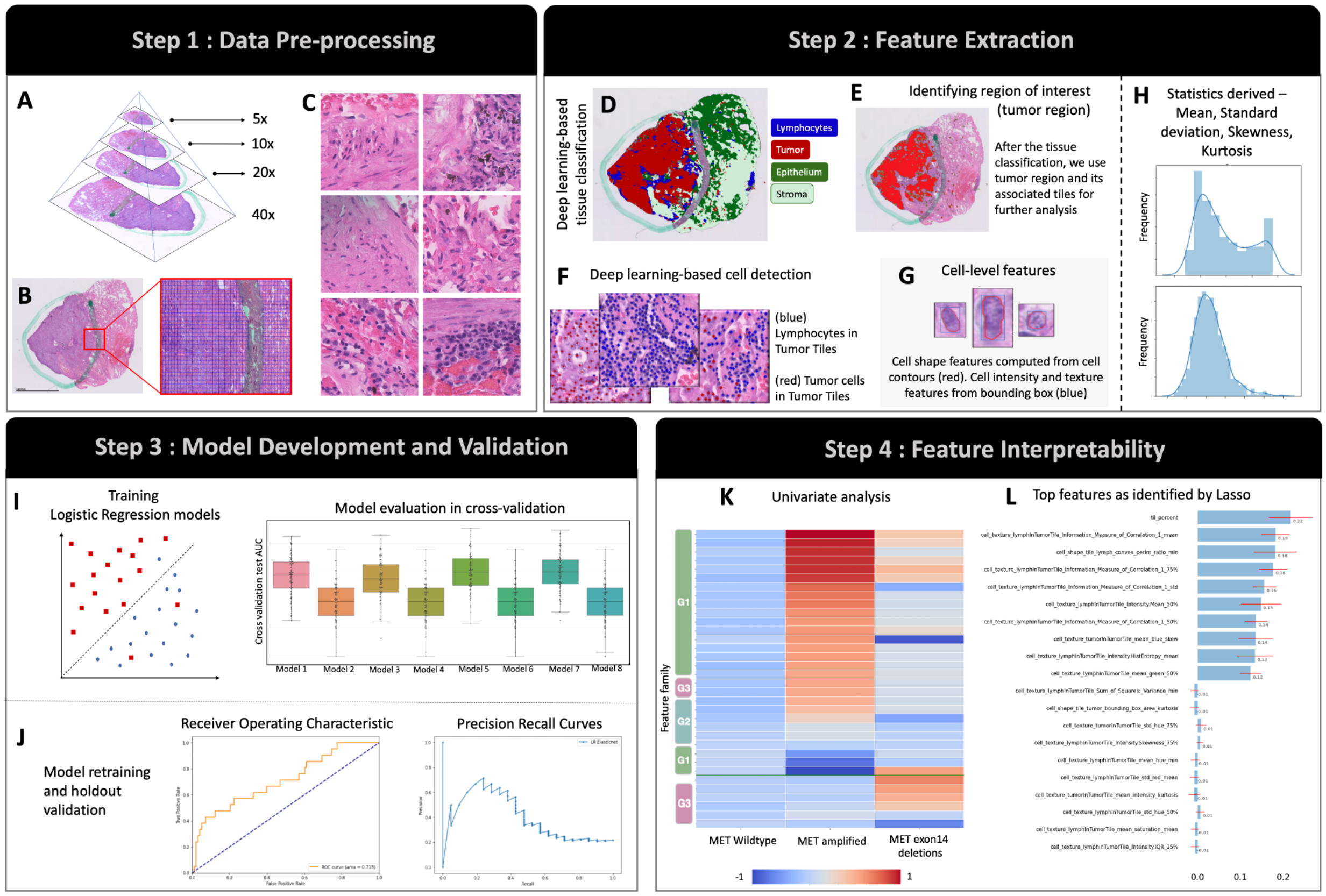

**Figure 1**: Overview of methods. (**A**) H&E stained images were scanned at 40X magnification and a 20X magnification of images was chosen for further analysis (**B**) An automated marker detection deep learning model was deployed to discard tiles with ink/marker on them. (**C**) Representative samples of 224x224 pixel tiles from the entire tissue image are displayed. (**D, E**) An automated deep learning segmentation model was deployed to identify the tumor region of the tissue. (**F**) A deep learning cell detection model was then deployed to identify tumor cells and tumor infiltrating lymphocytes within the tumor region. (**G, H**) Cell level morphological features, such as cell shape and cell bounding box texture were extracted and statistics were derived from the distribution of these statistics in the slide. (**I**) L1-penalized models were evaluated in a cross-validation setting to identify the best model hyperparameters. (**J**) The best identified model was re-trained on the entire training set and then evaluated on the independent holdout test set. (**K**) Univariate analysis identified features that were discriminative of *MET* wild-type from *MET* amplifications and *MET* exon 14 deletions, (**L**) The coefficients of the trained L1-penalized model were evaluated to identify important features that differentiate *MET* wild-type from *MET* altered tumors.

## Statistical analysis

The cohort was randomly shuffled and split into training and independent holdout test sets, stratified by target label (*MET* wild-type vs *MET* altered) and procedure type (Table 1).

For univariate analyses, *MET* wild-type (n = 425) vs *MET* amplified (n = 61) and *MET* wild-type (n = 425) vs *MET* exon 14 deletion (n = 39) distributions were compared using the Mann-Whitney U test. False discovery rate (FDR) correction with an alpha of 5%, was applied to generate q-values (Benjamini and Hochberg 1995). A threshold of 0.05 was applied to identify significant features from the two comparisons. For visualization purposes, each of the significant features was then normalized by subtracting its median and dividing by its median absolute deviation. Next, each group of interest (*MET* wild-type, amplified or exon 14 deletion), was summarized using the group's median value and plotted as a representative heatmap of features that were statistically significant in differentiating *MET* wild-type from *MET* alterations.

For supervised model development, a feature transformation pipeline was built using the scikit-learn (v1.0.2) framework in Python. First, count-based features (such as the total number of lymphocytes to percentage of total cells) were calculated. Next, area-based mean-summary features and standard deviation summary features were log transformed. Features were then mean-centered and divided by a robust measure of standard deviation using the median absolute deviation (features with few unique values were removed). Finally, using this base feature transformation pipeline, models were trained in nested cross-validation on the training set, including L1-penalized logistic regression models and relaxed L1-penalized logistic regression models. Samples with *MET* amplified and *MET* exon 14 deletions were joined into a single *MET* altered group due to limited numbers of *MET* altered cases. As a result, the supervised model target was a binary value of *MET* wild-type vs *MET* altered. In the inner loop of cross-validation, L1-penalization weight was tuned; for the relaxed L1-penalized logistic regression model, an additional penalization weight was tuned. In the outer loop of cross-validation, the generalization performance of L1-penalized and relaxed L1-penalized logistic regression models applying different class weighting ratios was compared. The "balanced" mode of class weights uses the labels *y* to automatically adjust weights inversely proportional to class frequencies in the input data as *n_samples / [n_classes * np.bincount(y)]*. Each cross-validation loop contained 10 repeats of 10 folds. The model and class weighting with the highest generalization performance as measured by ROC-AUC was selected, which was an L1-penalized model with a class weight ratio of 0.3 to 0.7 for *MET* wild-type to *MET* altered. This model was retrained on the entire training set, re-tuning the penalization weight using the inner cross-validation loop. Holdout performance was determined by performing inference with this retrained model on the independent holdout test set.

## Results

### Cohort characteristics

WSIs from NSCLC lung tissue along with associated NGS-based labels for *MET* amplification and exon 14 deletion were assembled from Tempus's clinicogenomic database.

A detailed list of inclusion and exclusion criteria is available in the methods. From a subset of histologic H&E-stained images, 704 NSCLC patients met the inclusion criteria (Table 1); this included 138 cases with *MET* amplification (CN>=5) or exon 14 deletion and 566 *MET* wild-type cases (no evidence of exon 14 deletion and NGS showing CN=2). This cohort was split into training (n=528) and holdout (n=176) sets using stratification on *MET* wild-type versus *MET* altered label and procedure type. No significant associations were found between NSCLC histological subtype or procedure type used to acquire patient tissue and *MET* labels (chi-square test, p>0.05).

### Univariate analysis establishes presence of unique imaging-based phenotype for different *MET* alterations

We hypothesized that a lymphocyte- and tumor-cell morphological signal exists within areas of NSCLC tumors that is related to *MET* alteration status. We developed a cell-based feature extraction pipeline to quantify shape, color, intensity, texture, and cell type percentage of cells (Figure 1, see Methods for details). A summary of extracted features is available in Supplementary Table 1. Using a univariate analysis, we investigated features that significantly differentiate *MET* wild-type from *MET* amplifications or *MET* exon 14 deletions (Figure 2). Of the 372 features investigated, we identified a total of 48 features as significantly associated with either *MET* amplification or exon 14 deletion (q-value < 0.05, FDR = 5%, Mann-Whitney U test) within the training set (see Supplementary Table 2 for a full list).

A total of 13 features were found to differentiate *MET* wild-type tumors from those harboring *MET* amplifications, with higher feature values found in the *MET* amplified setting. Of these features, 12/13 were related to the presence of tumor infiltrating lymphocytes (TILs) and 9/12 were specifically related to TIL cell-shape. Most features were related to increased skewness or kurtosis of TIL cell shape features in the *MET* amplified setting compared to *MET* wild-type, suggesting increased TIL pleomorphism for *MET* amplified tumors. Further, 37 features were identified that distinguish *MET* wild-type from *MET* exon 14 deletions, including 21 features that were decreased and 16 features that were increased in cases of *MET* exon 14 deletion. Both tumor cellular and TIL features were well represented. 12/14 increased features and 15/21 decreased features were related to pixel value information (either color-based or grayscale intensity-based); the remaining features were texture (8) or shape-based (2). Interestingly, tumor cell mean grayscale intensity was higher in *MET* exon 14 than *MET* wild-type, suggesting darker nuclei in *MET* exon 14 cases. Two features identified in the univariate analysis discriminated *MET* wild-type from both *MET* amplified and *MET* exon 14 deletions and these were related to pixel value information: the mean of hue of tumor cells and the mean of hue of lymphocytes.

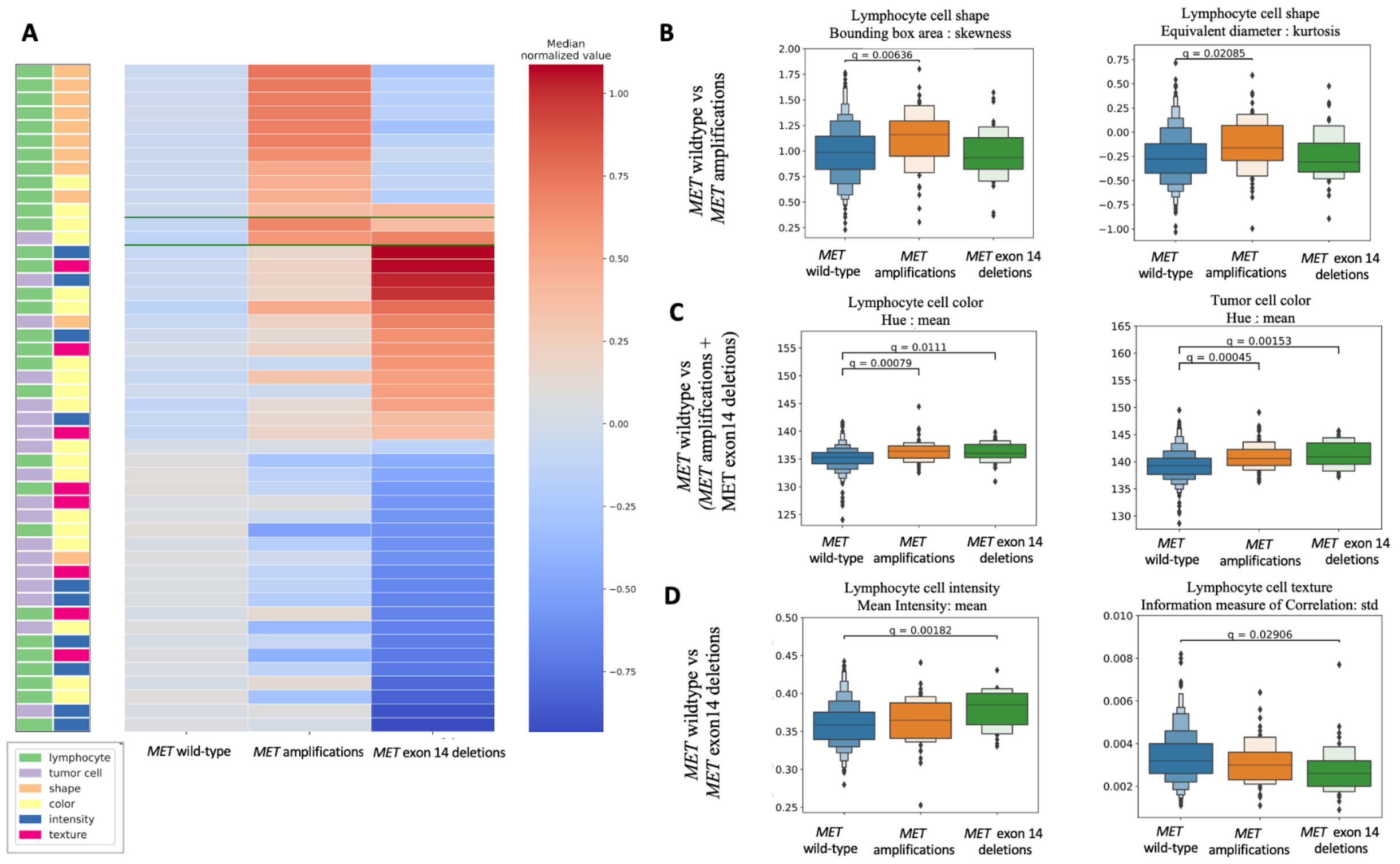

**Figure 2. Feature selection by univariate nonparametric statistical test to distinguish MET altered NSCLC tumors from MET wild-type. (A)** Nonparametric Mann-Whitney U test was used to measure statistical difference between feature distributions of samples with different *MET* alterations. Two comparisons were made: *MET* wild-type vs *MET* amplified and *MET* wild-type vs *MET* exon 14 deletions. FDR correction with alpha = 0.05 was used to obtain corrected values (q-values). Only features with $q < 0.05$ were selected. Two horizontal green lines separate features that were found to significantly distinguish *MET* wild-type vs amplified (above the top line), wild-type vs exon 14 deletion (below the bottom line), and distinguish *MET* wild-type from both *MET* amplified and *MET* exon 14 deletion (in between the two lines). For each group the features were sorted by their median normalized value, indicated by colorbar to the right. The colorbar to the left contains two columns; the first indicates cell type (lymphocyte or tumor), while the second indicates feature family (shape, color, grayscale intensity, or texture). **(B)** Boxen plots illustrate two of the Tumor infiltrating lymphocyte (TILs) shape-based features that were discriminative of *MET* wild-type from *MET* amplifications. These features were not found to be statistically significant in discriminating *MET* wild-type from *MET* exon 14 deletion NSCLC cases. For visual clarity, one outlier in the *MET* exon 14 group was cropped. **(C)** Boxen plots illustrate the two color features of TILs and tumor cells that were discriminative of *MET* wild-type from *MET* amplifications and *MET* exon 14 deletions. **(D)** Boxen plots illustrate intensity and texture features of TILs that were discriminative of *MET* wild-type from *MET* exon 14 deletions. These features were not found to be statistically significant in discriminating *MET* wild-type from *MET* amplifications.

## Computational imaging-based features discriminate *MET* wild-type NSCLC cases from *MET* alterations

The identification of discriminative cellular features in our univariate analysis suggested that we could develop a classifier to predict the presence of *MET* alterations from H&E WSIs. Given a desire for model interpretability and limited numbers of cases within *MET* amplification and *MET* exon 14 classes, we chose to create a binary logistic classification model with a sparse set of features. This model was used to predict *MET* wild-type versus *MET* altered status. Using the training set, we compared relaxed and non-relaxed versions of least absolute shrinkage and selection operator (LASSO) logistic regression models in nested cross-validation (see Methods). A non-relaxed LASSO model yielded the highest ROC-AUC of 0.77 ± 0.05 in the 10 repeats of 10-fold cross-validation (Figure 3A).

To identify consistently selected features from our dataset, we inspected the features selected by the LASSO model across cross-validation folds that have an interquartile range of coefficients below zero (indicating consistent association with *MET* wild-type) or above zero (indicating consistent association with *MET* altered status). 19 features met these criteria, including 12 features associated with TILs and 7 features associated with tumor cells (Figure 3B). Of the 12 features associated with TILs, 1 feature captured the percentage of TILs, 6 features investigated the variation in RGB color channels and grayscale intensity, and the remaining 5 features evaluated the textural attributes of the TILs using Haralick and Gradient features. Similarly, of the 7 features associated with tumor cells, 4 features captured the color variations and the remaining two investigated the Haralick texture-based differences.

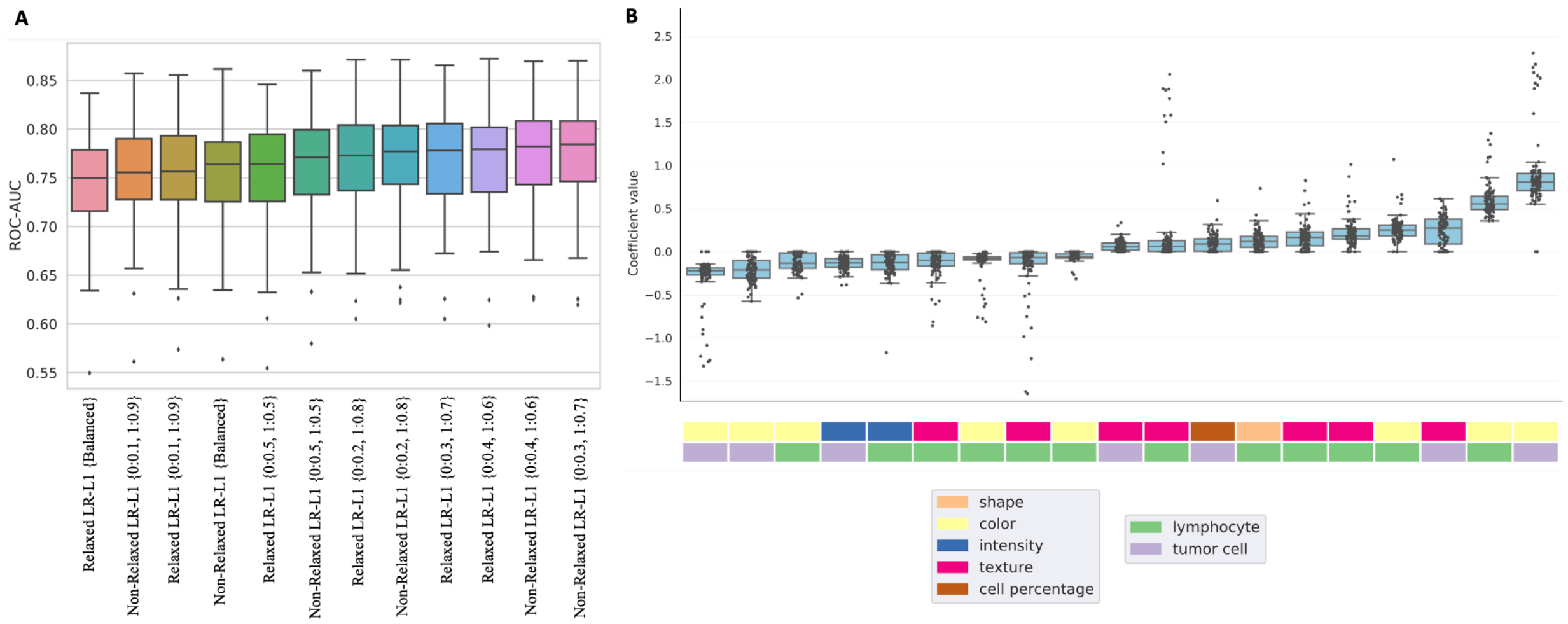

**Figure 3. Cross-validation evaluation and feature interpretability**. **(A)** Comparison by ROC-AUC of relaxed and non-relaxed LASSO models with different class-weights in the loss function. The best performing model was a non-relaxed LASSO model with an ROC-AUC of 0.77 ± 0.05 in the 10 repeats of 10-fold cross-validation. The *"balanced"* mode of class weights automatically adjusts weights inversely proportional to class frequencies in the input data. **(B)** The model coefficients of the best performing model were investigated to gain insight into the features that had good predictive power to discriminate *MET* wild-type from *MET* altered tumors, where 19 features were identified, including 12 features associated with TILs and 7 features associated with tumor cells.

Next, we re-trained the selected LASSO model on the entire training cohort and evaluated the model on the independent holdout test set. On the independent holdout set, the model achieved an ROC-AUC of 0.77 and an average precision of 0.56 (Figure 4A, 4B). The features consistently selected in cross-validation were a subset of features selected after re-training on this larger dataset. The spearman rank-correlation of coefficient values of this feature-subset was 0.97, suggesting the model training process was stable to underlying data.

After re-training, the classifier selected 43 features, which included 25 features associated with TILs and 18 features associated with tumor cells (Figure 4C, Supplementary Table 3). TIL and tumor feature coefficients were both negative (indicating association with *MET* wild-type) and positive (indicating association with *MET* altered state). Of the 25 features associated with TILs, 6 features quantified the shape of TILs, 11 features captured the variation in RGB color channels and grayscale intensity, and the remaining 8 features evaluated the textural attributes of the TILs using Haralick and Gradient features. Similarly, of the 18 features related to tumor cells, 7 investigated the textural attributes of the TILs using Haralick and Gradient features, 7 additional features investigated the variation in RGB color channels and grayscale intensity, and the remaining 4 features quantified the shape or percentage of tumor cells. An increased percentage of tumor cells relative to TILs within the tumor region was associated with *MET* altered status.

Furthermore, we also investigated the model's ability to differentiate *MET* wild-type (n=141) from *MET* amplifications (n=15) and *MET* exon 14 deletions (n=20) separately on the independent holdout test set. We found that *MET* wild-type cases differentiate from *MET* amplification tumors with an improved ROC-AUC of 0.89, whereas the *MET* wild-type cases differentiated from *MET* exon 14 deletions with an ROC-AUC of 0.66 (Figure 4A).

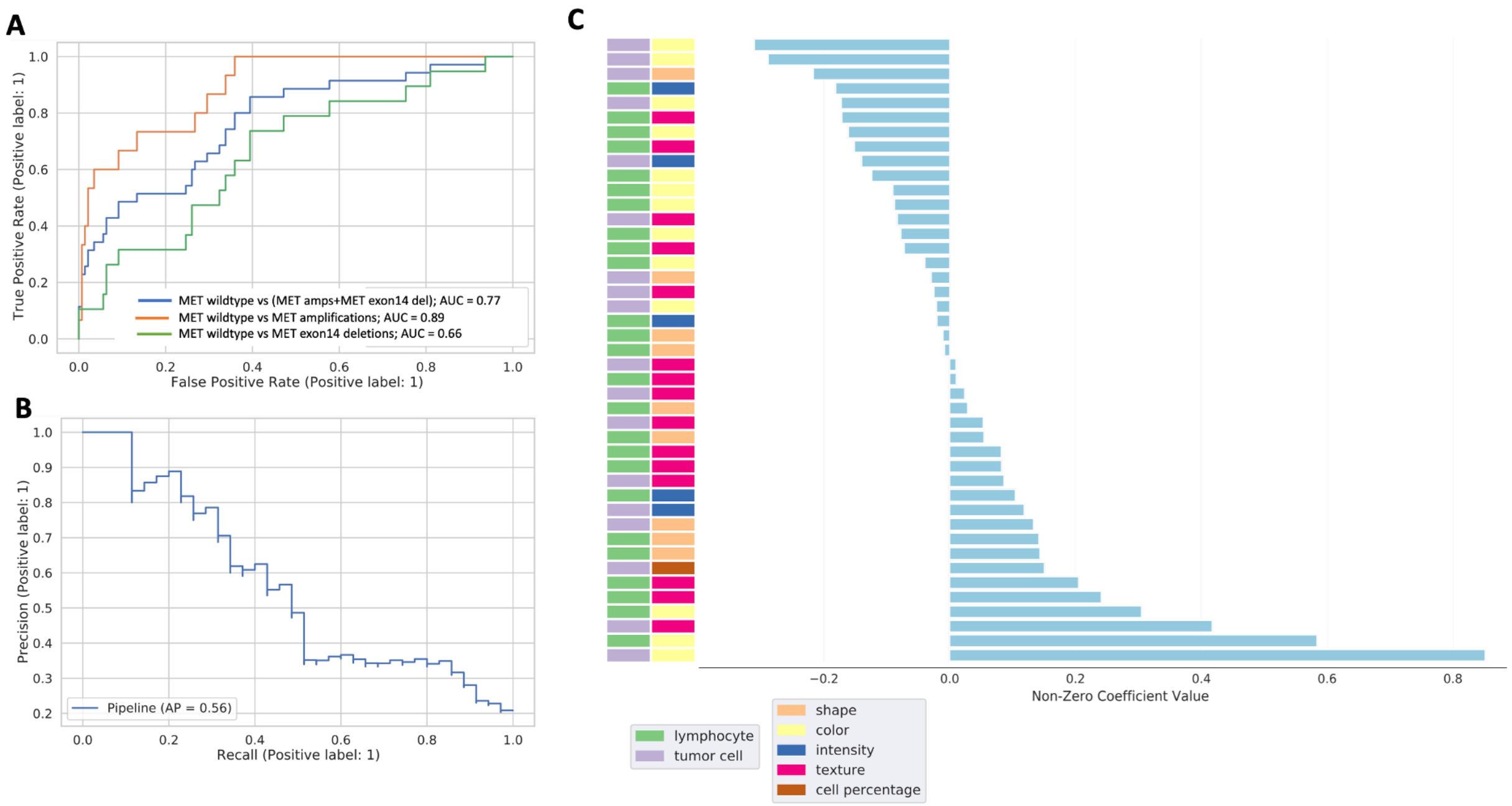

**Figure 4. Independent Holdout test set evaluation and feature interpretability**. **(A)** The best performing model was retrained on the entire training cohort and evaluated on the independent independent holdout test set (ROC-AUC = 0.77 ± 0.05) **(B)** The Precision-Recall curve with an average precision of 0.56 was obtained for the same model on the independent holdout test set **(C)** The model coefficients of the trained model were investigated to gain insight into the features that had good predictive power to discriminate *MET* wild-type from *MET* altered tumors in a supervised manner;the model identified 43 features that included 25 features associated with TILs and 18 features associated with tumor cells.

## Discussion

*MET* is a proto-oncogenic tyrosine kinase whose activation in NSCLC leads to uncontrolled cell growth, tumor formation, and tumor progression (Liang and Wang 2020; Organ and Tsao 2011; Zhang et al. 2018). *MET* is an attractive target for therapeutic inhibition (Socinski, Pennell, and Davies 2021; Zhang et al. 2018). While *MET* alterations can be assayed through a combination of FISH and NGS, we hypothesized that *MET* alteration additionally leads to changes in tumor and lymphocyte morphology that are detectable using computational methods. *MET* alterations fall predominantly into two categories: *MET* exon 14 deletions and *MET* amplifications. Drugs targeting *MET* exon 14 deletion have recently received FDA approval, while drugs targeting *MET* amplifications have shown promising initial results in ongoing clinical trials (Camidge et al. 2021; Wolf et al. 2020). Our analysis identified computational features that differentiate *MET* wild-type from either class of *MET* alterations. We further demonstrated that an L1-penalized model based on images of tumor infiltrating lymphocytes and tumor cells was predictive of *MET* alterations.

Previous studies have demonstrated an association between *MET* amplification and TILs. (Yoshimura et al. 2020). Similarly, the number of CD8+ TILs has been reported to be significantly increased in tumors with *MET* overexpression (Koh et al. 2015). In our univariate analysis, cell-shape features of the TILs were identified as the most commonly represented features that differentiate *MET* wild-type tumors from those harboring *MET* amplifications, with higher mean values in the *MET* wild-type compared to *MET* amplifications. In addition, cell-shape TIL features were predictive of *MET* alteration status in our logistic regression model, in both cross-validation and when re-trained on the entire training dataset. Future investigation of the biological significance of TIL cell shape in the setting of *MET* amplification and whether TIL cell-shape changes are associated with more suitable candidates for immune checkpoint inhibitors may be a promising area for future research.

Features that distinguished *MET* wild-type from *MET* exon 14 deletions in the univariate analysis were derived from several feature groups (shape, color, grayscale intensity, and texture) and from both TILs and tumor cells—unlike the consistent TIL shape features discovered to discriminate cases of *MET* amplification. The increased number and variety of significant features in this comparison could be due to differences in sample size or other clinical characteristics of these groups. Darker nuclei were a feature of *MET* exon 14 cases, and we hypothesize that *MET* exon 14 cases may have increased condensed chromatin compared to *MET* wild-type. Seven texture features from the Haralick family significantly differentiated *MET* wild-type from *MET* exon 14 deletions. Haralick features have been useful for predicting survival outcomes in NSCLC, suggesting their importance as a computational feature descriptor in cancer (Yu et al. 2016). A previous study has suggested an association between *MET* exon 14 deletions and clear cell features, hyaline globules and nuclear pleomorphism (Hayashi et al. 2021). Future work with larger datasets and rich clinical covariate information would enable us to verify the relationship between histological features identified previously, and further investigate the computational features found in this study as well.

Although our study is underpowered to detect significant differences in features between *MET* exon 14 and *MET* amplified cases, our univariate analysis suggests interesting trends. The same TIL cell shape features that differentiate *MET* amplified from *MET* wild-type were similar between *MET* exon 14 and *MET* wild-type. In addition, certain TIL RGB color channel and grayscale intensity features that differentiate *MET* exon 14 and *MET* wild-type were similar between *MET* amplified and *MET* wild-type. We hypothesize that these TIL features may distinguish *MET* amplified and *MET* exon 14 cases and examination of the relationship of these features to underlying biology would be interesting.

In our supervised model predicting *MET* alterations, we found a wide variety of selected features in both cross-validation as well as in the coefficients of the final model retrained on the full training set. Reassuringly, all features that were consistently predictive in cross-validation were selected when retraining on the full training set, demonstrating the robustness of these features and our modeling framework. While some feature overlap was found with the univariate analysis, the predictive model also yielded some differences. A direct comparison of feature importance between the univariate analysis and predictive model is difficult due to correlation of features within the feature space and our use of L1-penalization to yield a sparse feature set (Zou 2006). In the setting of a strongly correlated set of features, L1-penalization will choose a single feature in the set even if the other features are also predictive of the class label. Qualitatively, we found that important features were of similar class (color, intensity, shape, texture) between the two analyses, with the predictive model selecting more skewness and kurtosis-based features than the univariate analysis.

Interestingly, our supervised model performed substantially better on independent holdout test *MET* amplified samples (ROC-AUC = 0.89) than *MET* exon 14 samples (ROC-AUC = 0.66). This may be reflective of the larger number and diversity of *MET* exon 14 discriminative features seen in the univariate analysis, such that an easier to learn and simpler decision boundary exists to separate wild-type from *MET* amplified than *MET* exon 14 cases. Alternatively or simultaneously, the lower representation of *MET* exon 14 samples in the training data may bias the model to predict better on amplified cases. Given the differences in distinguishing features in the univariate analysis and model performance in the logistic regression model, development of softmax regression models to distinguish between *MET* wild-type, *MET* amplifications and *MET* exon 14 deletions using a larger dataset could prove beneficial to predictive performance.

The retrospective study presented here has several limitations that can be addressed by future studies. Although sample cases at Tempus Labs are received from diverse sources, validation of the results presented here on external sources would further support our findings. We did not explicitly explore differences in cellular features between *MET* exon 14 and *MET* amplified cases or develop predictive models to distinguish these classes, avenues that would be facilitated by larger datasets. In addition, various clinical covariates including stage, smoking status, patient sex, or prior therapy may play a confounding role in some of the imaging features we identified. In particular, *MET* amplification is a known resistance mechanism to first- or second-generation EGFR tyrosine kinase inhibitors including osimertinib (Leonetti et al. 2019). Future work could focus on confirming our findings while accounting for potential confounders, with a specific focus on prior therapy in NSCLC *EGFR* mutant tumors. In order to prevent overfitting of our predictive model, we used a restricted feature space. Other possible features such as spatial graphs, gradient, and texture entropy features may also capture morphological differences that distinguish *MET* wild-type samples from *MET* amplifications. Lastly, while this work used hand-crafted features derived from cells detected by deep learning, future approaches may also find improved accuracy and detection of *MET* altered NSCLC tumors using an end-to-end deep learning-based approach.

To our knowledge, this is the first study to predict *MET* alteration status in NSCLC using H&E WSIs. We found that computational imaging-based histological features from both TILs and tumor cells are important for explaining predictions, indicating the presence of a detectable morphological signal in tumor cells and lymphocytes. The proof-of-concept described here could facilitate the development of low-cost predictive models based on H&E-stained WSIs to improve screening for *MET* altered tumors as well as open up new avenues for further clinical research.

## References

Althammer, Sonja, Tze Heng Tan, Andreas Spitzmüller, Lorenz Rognoni, Tobias Wiestler, Thomas Herz, Moritz Widmaier, et al. 2019. “Automated Image Analysis of NSCLC Biopsies to Predict Response to Anti-PD-L1 Therapy.” *Journal for ImmunoTherapy of Cancer* 7 (1): 121. https://doi.org/10.1186/s40425-019-0589-x.

Alvarez-Jimenez, Charlems, Alvaro A. Sandino, Prateek Prasanna, Amit Gupta, Satish E. Viswanath, and Eduardo Romero. 2020. “Identifying Cross-Scale Associations between Radiomic and Pathomic Signatures of Non-Small Cell Lung Cancer Subtypes: Preliminary Results.” *Cancers* 12 (12): 3663. https://doi.org/10.3390/cancers12123663.

Awad, Mark M., Geoffrey R. Oxnard, David M. Jackman, Daniel O. Savukoski, Dimity Hall, Priyanka Shivdasani, Jennifer C. Heng, et al. 2016. “MET Exon 14 Mutations in Non-Small-Cell Lung Cancer Are Associated With Advanced Age and Stage-Dependent MET Genomic Amplification and c-Met Overexpression.” *Journal of Clinical Oncology: Official Journal of the American Society of Clinical Oncology* 34 (7): 721–30. https://doi.org/10.1200/JCO.2015.63.4600.

Bankhead, Peter, Maurice B. Loughrey, José A. Fernández, Yvonne Dombrowski, Darragh G. McArt, Philip D. Dunne, Stephen McQuaid, et al. 2017. “QuPath: Open Source Software for Digital Pathology Image Analysis.” *Scientific Reports* 7 (1): 16878. https://doi.org/10.1038/s41598-017-17204-5.

Beaubier, Nike, Robert Tell, Robert Huether, Martin Bontrager, Stephen Bush, Jerod Parsons, Kaanan Shah, et al. 2018. “Clinical Validation of the Tempus XO Assay.” *Oncotarget* 9 (40): 25826. https://doi.org/10.18632/oncotarget.25381.

Beaubier, Nike, Robert Tell, Denise Lau, Jerod R. Parsons, Stephen Bush, Jason Perera, Shelly Sorrells, et al. 2019. “Clinical Validation of the Tempus XT Next-Generation Targeted Oncology Sequencing Assay.” *Oncotarget* 10 (24): 2384. https://doi.org/10.18632/oncotarget.26797.

Benjamini, Yoav, and Yosef Hochberg. 1995. “Controlling the False Discovery Rate: A Practical and Powerful Approach to Multiple Testing.” *Journal of the Royal Statistical Society. Series B (Methodological)* 57 (1): 289–300.

Bera, Kaustav, Kurt A. Schalper, David L. Rimm, Vamsidhar Velcheti, and Anant Madabhushi. 2019. “Artificial Intelligence in Digital Pathology — New Tools for Diagnosis and Precision Oncology.” *Nature Reviews. Clinical Oncology* 16 (11): 703. https://doi.org/10.1038/s41571-019-0252-y.

Bradski, G. 2000. “The OpenCV Library.” *Dr. Dobb’s Journal of Software Tools*. http://citebay.com/how-to-cite/opencv/.

Bussolati, Gianni, Caterina Marchiò, Laura Gaetano, Rosanna Lupo, and Anna Sapino. 2008. “Pleomorphism of the Nuclear Envelope in Breast Cancer: A New Approach to an Old Problem.” *Journal of Cellular and Molecular Medicine* 12 (1): 209–18. https://doi.org/10.1111/j.1582-4934.2007.00176.x.

Camidge, D. Ross, Gregory A. Otterson, Jeffrey W. Clark, Sai-Hong Ignatius Ou, Jared Weiss, Steven Ades, Geoffrey I. Shapiro, et al. 2021. “Crizotinib in Patients With MET-Amplified NSCLC.” *Journal of Thoracic Oncology* 16 (6): 1017–29. https://doi.org/10.1016/j.jtho.2021.02.010.

Cheng, Y., J. Zhou, S. Lu, Y. Zhang, J. Zhao, H. Pan, Y.-M. Chen, et al. 2018. “Phase II Study of Tepotinib + Gefitinib (TEP+GEF) in MET-Positive (MET+)/Epidermal Growth Factor Receptor (EGFR)-Mutant (MT) Non-Small Cell Lung Cancer (NSCLC).” *Annals of Oncology* 29 (October): viii493. https://doi.org/10.1093/annonc/mdy292.

Coelho, Luis Pedro. 2013. “Mahotas: Open Source Software for Scriptable Computer Vision.” *Journal of Open Research Software*, no. 1(1): p.e3. https://doi.org/10.5334/jors.ac.

Corredor, Germán, Xiangxue Wang, Yu Zhou, Cheng Lu, Pingfu Fu, Konstantinos Syrigos, David L. Rimm, et al. 2019. “Spatial Architecture and Arrangement of Tumor-Infiltrating Lymphocytes for Predicting Likelihood of Recurrence in Early-Stage Non-Small Cell Lung Cancer.” *Clinical Cancer Research: An Official Journal of the American Association for Cancer Research* 25 (5): 1526–34. https://doi.org/10.1158/1078-0432.CCR-18-2013.

Coudray, Nicolas, Paolo Santiago Ocampo, Theodore Sakellaropoulos, Navneet Narula, Matija Snuderl, David Fenyö, Andre L. Moreira, Narges Razavian, and Aristotelis Tsirigos. 2018. “Classification and Mutation Prediction from Non–Small Cell Lung Cancer Histopathology Images Using Deep Learning.” *Nature Medicine* 24 (10): 1559–67. https://doi.org/10.1038/s41591-018-0177-5.

Diao, James A., Jason K. Wang, Wan Fung Chui, Victoria Mountain, Sai Chowdary Gullapally, Ramprakash Srinivasan, Richard N. Mitchell, et al. 2021. “Human-Interpretable Image Features Derived from Densely Mapped Cancer Pathology Slides Predict Diverse Molecular Phenotypes.” *Nature Communications* 12 (1): 1613. https://doi.org/10.1038/s41467-021-21896-9.

Dice, Lee R. 1945. “Measures of the Amount of Ecologic Association Between Species.” *Ecology* 26 (3): 297–302. https://doi.org/10.2307/1932409.

Drilon, Alexander, Federico Cappuzzo, Sai-Hong Ignatius Ou, and D. Ross Camidge. 2017. “Targeting MET in Lung Cancer: Will Expectations Finally Be MET?” *Journal of Thoracic Oncology: Official Publication of the International Association for the Study of Lung Cancer* 12 (1): 15–26. https://doi.org/10.1016/j.jtho.2016.10.014.

Fang, Lianghua, Hui Chen, Zhenya Tang, Neda Kalhor, Ching-Hua Liu, Hui Yao, Shimin Hu, et al. 2018. “MET Amplification Assessed Using Optimized FISH Reporting Criteria Predicts Early Distant Metastasis in Patients with Non-Small Cell Lung Cancer.” *Oncotarget* 9 (16): 12959. https://doi.org/10.18632/oncotarget.24430.

Haralick, R.M. 1979. “Statistical and Structural Approaches to Texture.” *Proceedings of the IEEE* 67 (5): 786–804. https://doi.org/10.1109/PROC.1979.11328.

Haralick, Robert M., K. Shanmugam, and Its’Hak Dinstein. 1973. “Textural Features for Image Classification.” *IEEE Transactions on Systems, Man, and Cybernetics* SMC-3 (6): 610–21. https://doi.org/10.1109/TSMC.1973.4309314.

Hayashi, Takuo, Shinji Kohsaka, Kazuya Takamochi, Satsuki Kishikawa, Daiki Ikarashi, Kei Sano, Kieko Hara, et al. 2021. “Histological Characteristics of Lung Adenocarcinoma with Uncommon Actionable Alterations: Special Emphasis on MET Exon 14 Skipping Alterations.” *Histopathology* 78 (7): 987–99. https://doi.org/10.1111/his.14311.

Heydt, Carina, Ann-Kathrin Becher, Svenja Wagener-Ryczek, Markus Ball, Anne M. Schultheis, Simon Schallenberg, Vanessa Rüsseler, Reinhard Büttner, and Sabine Merkelbach-Bruse. 2019. “Comparison of in Situ and Extraction-Based Methods for the Detection of MET Amplifications in Solid Tumors.” *Computational and Structural Biotechnology Journal* 17 (January): 1339–47. https://doi.org/10.1016/j.csbj.2019.09.003.

Hu, Jun, Jerod Parsons, Brittany Mineo, Josh SK Bell, Jenna Malinauskas, Joshua Drews, Jack Michuda, et al. 2021. “Abstract 2239: Comprehensive Validation of RNA Sequencing for Clinical NGS Fusion Genes and RNA Expression Reporting.” *Cancer Research* 81 (13_Supplement): 2239. https://doi.org/10.1158/1538-7445.AM2021-2239.

Jevtić, Predrag, Lisa J. Edens, Lidija D. Vuković, and Daniel L. Levy. 2014. “Sizing and Shaping the Nucleus: Mechanisms and Significance.” *Current Opinion in Cell Biology* 0 (June): 16. https://doi.org/10.1016/j.ceb.2014.01.003.

Jiang, Wenxiao, Guiqing Cai, Peter C. Hu, and Yue Wang. 2018. “Personalized Medicine in Non-Small Cell Lung Cancer: A Review from a Pharmacogenomics Perspective.” *Acta Pharmaceutica Sinica. B* 8 (4): 530–38. https://doi.org/10.1016/j.apsb.2018.04.005.

Koh, Jaemoon, Heounjeong Go, Bhumsuk Keam, Moon-Young Kim, Soo Jeong Nam, Tae Min

Kim, Se-Hoon Lee, et al. 2015. “Clinicopathologic Analysis of Programmed Cell Death-1 and Programmed Cell Death-Ligand 1 and 2 Expressions in Pulmonary Adenocarcinoma: Comparison with Histology and Driver Oncogenic Alteration Status.” *Modern Pathology* 28 (9): 1154–66. https://doi.org/10.1038/modpathol.2015.63.

Leonetti, Alessandro, Sugandhi Sharma, Roberta Minari, Paola Perego, Elisa Giovannetti, and Marcello Tiseo. 2019. “Resistance Mechanisms to Osimertinib in EGFR-Mutated Non-Small Cell Lung Cancer.” *British Journal of Cancer* 121 (9): 725–37. https://doi.org/10.1038/s41416-019-0573-8.

Li, Feng, Caichen Li, Xiuyu Cai, Zhanhong Xie, Liquan Zhou, Bo Cheng, Ran Zhong, et al. 2021. “The Association between CD8+ Tumor-Infiltrating Lymphocytes and the Clinical Outcome of Cancer Immunotherapy: A Systematic Review and Meta-Analysis.” *EClinicalMedicine* 41 (November). https://doi.org/10.1016/j.eclinm.2021.101134.

Liang, Hongge, and Mengzhao Wang. 2020. “MET Oncogene in Non-Small Cell Lung Cancer: Mechanism of MET Dysregulation and Agents Targeting the HGF/c-Met Axis.” *OncoTargets and Therapy* 13: 2491. https://doi.org/10.2147/OTT.S231257.

Lu, Cheng, David Romo-Bucheli, Xiangxue Wang, Andrew Janowczyk, Shridar Ganesan, Hannah Gilmore, David Rimm, and Anant Madabhushi. 2018. “Nuclear Shape and Orientation Features from H&E Images Predict Survival in Early-Stage Estrogen Receptor-Positive Breast Cancers.” *Laboratory Investigation; a Journal of Technical Methods and Pathology* 98 (11): 1438–48. https://doi.org/10.1038/s41374-018-0095-7.

Lynch, Thomas J., Daphne W. Bell, Raffaella Sordella, Sarada Gurubhagavatula, Ross A. Okimoto, Brian W. Brannigan, Patricia L. Harris, et al. 2004. “Activating Mutations in the Epidermal Growth Factor Receptor Underlying Responsiveness of Non-Small-Cell Lung Cancer to Gefitinib.” *The New England Journal of Medicine* 350 (21): 2129–39. https://doi.org/10.1056/NEJMoa040938.

Mathieu, Luckson N., Erin Larkins, Oladimeji Akinboro, Pourab Roy, Anup K. Amatya, Mallorie H. Fiero, Pallavi S. Mishra-Kalyani, et al. 2022. “FDA Approval Summary: Capmatinib and Tepotinib for the Treatment of Metastatic NSCLC Harboring MET Exon 14 Skipping Mutations or Alterations.” *Clinical Cancer Research* 28 (2): 249–54. https://doi.org/10.1158/1078-0432.CCR-21-1566.

Mukherjee, Richik N., Pan Chen, and Daniel L. Levy. 2016. “Recent Advances in Understanding Nuclear Size and Shape.” *Nucleus* 7 (2): 167. https://doi.org/10.1080/19491034.2016.1162933.

Organ, Shawna Leslie, and Ming-Sound Tsao. 2011. “An Overview of the C-MET Signaling Pathway.” *Therapeutic Advances in Medical Oncology* 3 (1 Suppl): S7. https://doi.org/10.1177/1758834011422556.

Park, Keunchil, Eric B. Haura, Natasha B. Leighl, Paul Mitchell, Catherine A. Shu, Nicolas Girard, Santiago Viteri, et al. 2021. “Amivantamab in EGFR Exon 20 Insertion–Mutated Non–Small-Cell Lung Cancer Progressing on Platinum Chemotherapy: Initial Results From the CHRYSALIS Phase I Study.” *Journal of Clinical Oncology* 39 (30): 3391–3402. https://doi.org/10.1200/JCO.21.00662.

Peng, Lun-Xi, Guang-Ling Jie, An-Na Li, Si-Yang Liu, Hao Sun, Mei-Mei Zheng, Jia-Ying Zhou, et al. 2021. “MET Amplification Identified by Next-Generation Sequencing and Its Clinical Relevance for MET Inhibitors.” *Experimental Hematology & Oncology* 10 (1): 52. https://doi.org/10.1186/s40164-021-00245-y.

Planchard, D., S. Popat, K. Kerr, S. Novello, E. F. Smit, C. Faivre-Finn, T. S. Mok, et al. 2018. “Metastatic Non-Small Cell Lung Cancer: ESMO Clinical Practice Guidelines for Diagnosis, Treatment and Follow-Up.” *Annals of Oncology: Official Journal of the European Society for Medical Oncology* 29 (Suppl 4): iv192–237. https://doi.org/10.1093/annonc/mdy275.

Ronneberger, Olaf, Philipp Fischer, and Thomas Brox. 2015. “U-Net: Convolutional Networks for Biomedical Image Segmentation.” *ArXiv:1505.04597 [Cs]*, May. http://arxiv.org/abs/1505.04597.
Sebastian V, Bino, A. Unnikrishnan, and Kannan Balakrishnan. 2012. “Gray Level Co-Occurrence Matrices: Generalisation and Some New Features.” *ArXiv:1205.4831 [Cs]*, May. http://arxiv.org/abs/1205.4831.
Sha, Lingdao, Boleslaw L. Osinski, Irvin Y. Ho, Timothy L. Tan, Caleb Willis, Hannah Weiss, Nike Beaubier, Brett M. Mahon, Tim J. Taxter, and Stephen S. F. Yip. 2019. “Multi-Field-of-View Deep Learning Model Predicts Nonsmall Cell Lung Cancer Programmed Death-Ligand 1 Status from Whole-Slide Hematoxylin and Eosin Images.” *Journal of Pathology Informatics* 10: 24. https://doi.org/10.4103/jpi.jpi_24_19.
Smits, Alexander J. J., J. Alain Kummer, Peter C. de Bruin, Mijke Bol, Jan G. van den Tweel, Kees A. Seldenrijk, Stefan M. Willems, et al. 2014. “The Estimation of Tumor Cell Percentage for Molecular Testing by Pathologists Is Not Accurate.” *Modern Pathology* 27 (2): 168–74. https://doi.org/10.1038/modpathol.2013.134.
Socinski, Mark A., Nathan A. Pennell, and Kurtis D. Davies. 2021. “MET Exon 14 Skipping Mutations in Non–Small-Cell Lung Cancer: An Overview of Biology, Clinical Outcomes, and Testing Considerations.” *JCO Precision Oncology*, no. 5 (November): 653–63. https://doi.org/10.1200/PO.20.00516.
Song, Zhengbo, Hong Wang, Zongyang Yu, Peihua Lu, Chunwei Xu, Gang Chen, and Yiping Zhang. 2019. “De Novo MET Amplification in Chinese Patients With Non-Small-Cell Lung Cancer and Treatment Efficacy With Crizotinib: A Multicenter Retrospective Study.” *Clinical Lung Cancer* 20 (2): e171–76. https://doi.org/10.1016/j.cllc.2018.11.007.
Sung, Hyuna, Jacques Ferlay, Rebecca L. Siegel, Mathieu Laversanne, Isabelle Soerjomataram, Ahmedin Jemal, and Freddie Bray. 2021. “Global Cancer Statistics 2020: GLOBOCAN Estimates of Incidence and Mortality Worldwide for 36 Cancers in 185 Countries.” *CA: A Cancer Journal for Clinicians* 71 (3): 209–49. https://doi.org/10.3322/caac.21660.
Tollis, Sylvain, Andrea Rizzotto, Nhan T. Pham, Sonja Koivukoski, Aishwarya Sivakumar, Steven Shave, Jan Wildenhain, et al. 2022. “Chemical Interrogation of Nuclear Size Identifies Compounds with Cancer Cell Line-Specific Effects on Migration and Invasion.” *ACS Chemical Biology*, February. https://doi.org/10.1021/acschembio.2c00004.
Wolf, Jürgen, Takashi Seto, Ji-Youn Han, Noemi Reguart, Edward B. Garon, Harry J.M. Groen, Daniel S.W. Tan, et al. 2020. “Capmatinib in MET Exon 14–Mutated or MET-Amplified Non–Small-Cell Lung Cancer.” *New England Journal of Medicine* 383 (10): 944–57. https://doi.org/10.1056/NEJMoa2002787.
Wu, Yi-Long, Li Zhang, Dong-Wan Kim, Xiaoqing Liu, Dae Ho Lee, James Chih-Hsin Yang, Myung-Ju Ahn, et al. 2018. “Phase Ib/II Study of Capmatinib (INC280) Plus Gefitinib After Failure of Epidermal Growth Factor Receptor (EGFR) Inhibitor Therapy in Patients With EGFR-Mutated, MET Factor–Dysregulated Non–Small-Cell Lung Cancer.” *Journal of Clinical Oncology* 36 (31): 3101–9. https://doi.org/10.1200/JCO.2018.77.7326.
Yang, Soo-Ryum, Anne M. Schultheis, Helena Yu, Diana Mandelker, Marc Ladanyi, and Reinhard Büttner. 2020. “Precision Medicine in Non-Small Cell Lung Cancer: Current Applications and Future Directions.” *Seminars in Cancer Biology*, July. https://doi.org/10.1016/j.semcancer.2020.07.009.
Yoshimura, Katsuhiro, Yusuke Inoue, Kazuo Tsuchiya, Masato Karayama, Hidetaka Yamada, Yuji Iwashita, Akikazu Kawase, et al. 2020. “Elucidation of the Relationships of MET Protein Expression and Gene Copy Number Status with PD-L1 Expression and the Immune Microenvironment in Non-Small Cell Lung Cancer.” *Lung Cancer (Amsterdam, Netherlands)* 141 (March): 21–31. https://doi.org/10.1016/j.lungcan.2020.01.005.
Yu, Kun-Hsing, Ce Zhang, Gerald J. Berry, Russ B. Altman, Christopher Ré, Daniel L. Rubin, and Michael Snyder. 2016. “Predicting Non-Small Cell Lung Cancer Prognosis by Fully

Automated Microscopic Pathology Image Features.” *Nature Communications* 7 (1): 12474. https://doi.org/10.1038/ncomms12474.

Zayed, Nourhan, and Heba A. Elnemr. 2015. “Statistical Analysis of Haralick Texture Features to Discriminate Lung Abnormalities.” *International Journal of Biomedical Imaging* 2015 (October): e267807. https://doi.org/10.1155/2015/267807.

Zhang, Yazhuo, Mengfang Xia, Ke Jin, Shufei Wang, Hang Wei, Chunmei Fan, Yingfen Wu, et al. 2018. “Function of the C-Met Receptor Tyrosine Kinase in Carcinogenesis and Associated Therapeutic Opportunities.” *Molecular Cancer* 17 (1): 45. https://doi.org/10.1186/s12943-018-0796-y.

Zou, Hui. 2006. “The Adaptive Lasso and Its Oracle Properties.” *Journal of the American Statistical Association* 101 (476): 1418–29. https://doi.org/10.1198/016214506000000735.

Zwillinger, Daniel, and Stephen Kokoska. 1999. *CRC Standard Probability and Statistics Tables and Formulae - AbeBooks* -.

# Acknowledgements

We thank Aïcha BenTaieb, Kim Blackwell, Adam Hockenberry, and Amrita Iyer for critical reading of this manuscript. We thank Debra Chin, Joshua S.K. Bell, Alessandra Breschi, Ryan Jones, Gaurav Khullar, Amy Welch for assistance developing the cohort used in this study. Finally, we thank Irvin Ho and members of the engineering team for development of the compute infrastructure used in this study.

# Supplementary

| Feature family | # features | Features Extracted | Description |
|---|---|---|---|
| Percentage | 4 | ● Tumor percent = number of tumor cells divided by number of cells<br>● TIL percent = number of lymphocytes within tumor tiles divided by number of tumor cells<br>● Lymphocyte percent - Percent of identified cells that are lymphocytes<br>● Non tumor lymphocyte percent - Percent of identified cells that are neither tumor cells nor lymphocytes | ● Percentages of indicated cells out of total number of detected cells |
| Shape of a given cell | 88 | ● Area,<br>● Bounding box area,<br>● Solidity (Contour Area/Convex Hull Area),<br>● Perimeter,<br>● Convex perimeter ratio (convex bull perimeter/perimeter),<br>● Circularity = area/perimeter^2,<br>● Aspect ratio = (Width/Height) of tumor bounding box ,<br>● Equivalent diameter = √4×ContourArea/π,<br>● Longest axis,<br>● Area divided by bounding box,<br>● Bounding box aspect ratio | ● 11 features describing shape of segmented cell nuclei (i.e. area, perimeter, and circularity).<br>● 4 derived summary statistics per feature.<br>● 2 cell types (lymphocyte & tumor) |
| Color | 96 | Mean and STD of:<br>● Red<br>● Green<br>● Blue<br>● Hue<br>● Saturation<br>● Intensity | ● 12 features describing RGB color of bounding boxes enclosing each detected cell nucleus<br>● 4 derived summary statistics per feature.<br>● 2 cell types: lymphocyte & tumor |
| Grayscale intensity | 56 | ● Min<br>● Max | ● 7 features describing grayscale intensity of |
| | | ● Mean<br>● Std<br>● IQR<br>● Skewness<br>● Kurtosis | bounding boxes enclosing each detected cell nucleus<br>● 4 derived summary statistics per feature.<br>● 2 cell types: lymphocyte & tumor |
| Texture | 128 | Haralik features (R.M. Haralick 1979; Zayed and Elnemr 2015; Sebastian V, Unnikrishnan, and Balakrishnan 2012):<br>● Angular Second Moment<br>● Contrast<br>● Correlation<br>● Sum of Squares: Variance,<br>● Inverse Difference Moment<br>● Sum Average<br>● Sum Variance<br>● Difference Variance<br>● Information Measure of Correlation 1<br>● Information Measure of Correlation 2<br><br>Statistics of Grayscale Gradient (Zwillinger and Kokoska 1999):<br>● Mean - Mean of gradient data<br>● Standard deviation - Standard deviation of gradient data<br>● Skewness - Skewness of gradient data.<br>● Kurtosis of gradient data - Kurtosis of gradient data<br>● Sum of canny filter - Sum of canny filtered gradient data<br>● Mean of canny filter | ● 16 features (10 Haralik, 6 Gradient) describing texture features of bounding boxes enclosing each detected cell nucleus<br>● 4 derived summary statistics per feature.<br>● 2 cell types: lymphocyte & tumor |
| **Total features extracted** | | **372** | |

**Table S1**: **List of features extracted.** A total of 372 features at 20X magnification were extracted per patient, corresponding to one of five categories: cell nuclei percentage, shape, RGB color, grayscale intensity, and texture.

| Cell of interest within tumor region | Feature Family | Feature | Statistic derived | Association w.r.t *MET* wild-type | Q-value *MET* wild-type vs *MET* amplifications | Q-value *MET* wild-type vs *MET* exon 14 deletions |
|---|---|---|---|---|---|---|
| *MET* wild-type vs *MET* amplifications | | | | | | |
| Lymphocytes | Shape | Bounding box area | Skewness | Increased | 0.006 | 0.994 |
| Lymphocytes | Shape | Length | Kurtosis | Increased | 0.004 | 0.989 |
| Lymphocytes | Shape | Length | Skewness | Increased | 0.016 | 0.994 |
| Lymphocytes | Shape | Perimeter | Skewness | Increased | 0.014 | 0.909 |
| Lymphocytes | Shape | Equivalent Diameter | Skewness | Increased | 0.005 | 0.972 |
| Lymphocytes | Shape | Area | Skewness | Increased | 0.002 | 0.994 |
| Lymphocytes | Shape | Area | Kurtosis | Increased | 0.002 | 0.968 |
| Lymphocytes | Shape | Bounding box area | Kurtosis | Increased | 0.022 | 0.989 |
| Lymphocytes | Color | Standard deviation of Hue | Standard deviation | Increased | 0.013 | 0.990 |
| Lymphocytes | Shape | Equivalent Diameter | Kurtosis | Increased | 0.021 | 0.972 |
| Lymphocytes | Color | Standard deviation of Hue | Mean | Increased | 0.006 | 0.202 |
| *MET* wild-type vs *MET* (amplifications + exon 14 deletions) | | | | | | |
| Lymphocytes | Color | Mean of Hue | Mean | Increased | 0.001 | 0.011 |
| Tumor cells | Color | Mean of Hue | Mean | Increased | <0.001 | 0.002 |
| *MET* wild-type vs *MET* exon 14 deletions | | | | | | |
| Lymphocytes | Grayscale intensity | Intensity mean | Mean | Increased | 0.478 | 0.002 |
| Lymphocytes | Texture | Haralick: Sum Average | Mean | Increased | 0.478 | 0.002 |
| Tumor cells | Grayscale | Minimum intensity | Mean | Increased | 0.574 | 0.003 |
| | intensity | | | | | |
| Lymphocytes | Color | Mean of Green Channel | Mean | Increased | 0.577 | 0.008 |
| Lymphocytes | Color | Mean of Red Channel | Mean | Increased | 0.128 | <0.001 |
| Tumor cells | Shape | Convex perimeter ratio | Skewness | Increased | 0.465 | 0.003 |
| Lymphocytes | Grayscale intensity | Minimum intensity | Mean | Increased | 0.566 | 0.011 |
| Lymphocytes | Texture | Haralick: Information measure of correlation 2 | Mean | Increased | 0.465 | 0.030 |
| Lymphocytes | Color | Mean intensity | Mean | Increased | 0.984 | 0.008 |
| Tumor cells | Color | Mean of Red Channel | Mean | Increased | 0.324 | 0.003 |
| Lymphocytes | Color | Mean of Blue Channel | Mean | Increased | 0.951 | 0.009 |
| Tumor cells | Color | Mean of Hue | Standard deviation | Increased | 0.438 | 0.015 |
| Tumor cells | Grayscale intensity | Mean intensity | Mean | Increased | 0.764 | 0.026 |
| Tumor cells | Texture | Haralick: Sum Average | Mean | Increased | 0.773 | 0.028 |
| Tumor cells | Color | Mean of Hue | Kurtosis | Decreased | 0.982 | 0.011 |
| Lymphocytes | Color | Mean of Red Channel | Skewness | Decreased | 0.275 | 0.019 |
| Tumor cells | Color | Mean of Hue | Skewness | Decreased | 0.489 | 0.009 |
| Lymphocytes | Texture | Haralick: Information measure of correlation 2 | Standard deviation | Decreased | 0.459 | 0.029 |
| Tumor cells | Texture | Haralick: Correlation | Kurtosis | Decreased | 0.835 | 0.030 |
| Tumor cells | Color | Mean of Green channel | Skewness | Decreased | 0.501 | 0.034 |
| Lymphocytes | Color | Standard deviation of Saturation | Standard deviation | Decreased | 0.100 | 0.036 |
| Tumor cells | Color | Standard deviation of Green Channel | Standard deviation | Decreased | 0.478 | 0.049 |
| Tumor cells | Shape | Convex perimeter ratio | Kurtosis | Decreased | 0.538 | 0.015 |
| Tumor cells | Texture | Haralick: Sum Average | Skewness | Decreased | 0.489 | 0.028 |
| Tumor cells | Grayscale intensity | Intensity mean | Skewness | Decreased | 0.512 | 0.026 |
| Tumor cells | Grayscale intensity | Minimum intensity | Kurtosis | Decreased | 0.305 | 0.008 |
| Lymphocytes | Texture | Gradient magnitude Skewness | Mean | Decreased | 0.984 | 0.049 |
| Tumor cells | Color | Standard deviation of Saturation | Standard deviation | Decreased | 0.218 | 0.009 |
| Lymphocytes | Grayscale intensity | Intensity skewness | Mean | Decreased | 0.478 | 0.012 |
| Lymphocytes | Texture | Haralick: Information measure of correlation 1 | Mean | Decreased | 0.305 | 0.030 |
| Lymphocytes | Grayscale intensity | Intensity kurtosis | Mean | Decreased | 0.478 | 0.008 |
| Lymphocytes | Color | Standard deviation of Blue Channel | Standard deviation | Decreased | 0.984 | 0.030 |
| Lymphocytes | Color | Mean of Saturation | Mean | Decreased | 0.478 | 0.036 |
| Tumor cells | Grayscale intensity | Maximum intensity | Standard deviation | Decreased | 0.776 | 0.008 |
| Lymphocytes | Grayscale intensity | Intensity kurtosis | Standard deviation | Decreased | 0.478 | 0.008 |

**Table S2**: Features identified in the univariate analysis. Nonparametric Mann-Whitney U test was used to measure statistical difference between feature distributions of samples with different *MET* alterations. Two comparisons were made: *MET* wild-type vs *MET* amplified and *MET* wild-type vs *MET* exon 14 deletions. FDR correction with alpha = 0.05 was used to obtain corrected values (q-values). Only features with $q < 0.05$ were selected.

| Cell of interest within tumor region | Feature Family | Feature | Statistic derived | Coefficient |
|---|---|---|---|---|
| Tumor cells | Color | Standard deviation of Saturation | Standard deviation | -0.311 |
| Tumor cells | Color | Mean Hue | Kurtosis | -0.289 |
| Tumor cells | Shape | Tumor circularity | Kurtosis | -0.217 |
| Lymphocytes | Intensity | Intensity skewness | Standard deviation | -0.181 |
| Tumor cells | Color | Mean of Blue channel | Mean | -0.172 |
| Lymphocytes | Texture | Standard deviation of Gradient Magnitude | Kurtosis | -0.172 |
| Lymphocytes | Color | Standard deviation of Saturation | Skewness | -0.161 |
| Lymphocytes | Texture | Haralick: Inverse Difference Moment | Standard deviation | -0.151 |
| Tumor cells | Intensity | Minimum of Intensity | Kurtosis | -0.140 |
| Lymphocytes | Color | Mean of Red channel | Skewness | -0.124 |
| Lymphocytes | Color | Mean of Blue channel | Standard deviation | -0.090 |
| Lymphocytes | Color | Mean of Blue channel | Kurtosis | -0.088 |
| Tumor cells | Texture | Harlick: Information Measure of Correlation 2 | Skewness | -0.083 |
| Lymphocytes | Color | Mean of Hue | Kurtosis | -0.077 |
| Lymphocytes | Texture | Harlick: Information Measure of Correlation 1 | Skewness | -0.072 |
| Lymphocytes | Color | Standard deviation of Saturation | Kurtosis | -0.039 |
| Tumor cells | Shape | Bounding box aspect ratio | Kurtosis | -0.029 |
| Tumor cells | Texture | Mean of Gradient Canny | Kurtosis | -0.025 |
| Tumor cells | Color | Standard deviation of Red channel | Skewness | -0.021 |
| Lymphocytes | Intensity | Intensity Kurtosis | Standard deviation | -0.020 |
| Lymphocytes | Shape | Aspect Ratio | Mean | -0.011 |
| Lymphocytes | Shape | Aspect Ratio | Kurtosis | -0.008 |
| Tumor cells | Texture | Mean of Gradient Magnitude | Kurtosis | 0.010 |
| Lymphocytes | Texture | Mean of Gradient Canny | Kurtosis | 0.010 |
| Tumor cells | Texture | Standard deviation of Gradient Magnitude | Kurtosis | 0.024 |
| Lymphocytes | Shape | Area over bounding box | Kurtosis | 0.029 |
| Tumor cells | Texture | Haralick: Contrast | Kurtosis | 0.053 |
| Lymphocytes | Shape | Solidity | Kurtosis | 0.054 |
| Lymphocytes | Texture | Mean of Gradient Magnitude | Kurtosis | 0.082 |
| Lymphocytes | Texture | Skewness of Gradient Magnitude | Kurtosis | 0.082 |
| Tumor cells | Texture | Inverse Difference Moment | Kurtosis | 0.086 |
| Lymphocytes | Intensity | Minimum Intensity | Kurtosis | 0.104 |
| Tumor cells | Intensity | Minimum Intensity | Standard deviation | 0.118 |
| Tumor cells | Shape | Convex perimeter ratio | Skewness | 0.133 |
| Lymphocytes | Shape | Length | Skewness | 0.142 |
| Lymphocytes | Shape | Area | Kurtosis | 0.143 |
| Tumor cells | Percent | Percentage | Percent | 0.150 |
| Lymphocytes | Texture | Haralick: Correlation | Kurtosis | 0.205 |
| Lymphocytes | Texture | Haralick: Information Measure of Correlation 1 | Kurtosis | 0.241 |
| Lymphocytes | Color | Standard deviation of Red channel | Kurtosis | 0.305 |
| Tumor cells | Texture | Kurtosis of Gradient Magnitude | Standard deviation | 0.417 |
| Lymphocytes | Color | Standard deviation of Hue | Mean | 0.584 |
| Tumor cells | Color | Mean of Hue | Mean | 0.851 |

**Table S3**: Lasso Model Coefficients when trained on the entire training cohort